\documentclass[letterpaper]{article} 
\usepackage{aaai2026} 
\usepackage{times} 
\usepackage{helvet} 
\usepackage{courier} 
\usepackage[hyphens]{url} 
\usepackage{graphicx} 
\usepackage{natbib} 
\usepackage{caption} 
\usepackage{algorithm}
\usepackage{algorithmic}

\usepackage{newfloat}
\usepackage{listings}
\DeclareCaptionStyle{ruled}{labelfont=normalfont,labelsep=colon,strut=off} 
\floatstyle{ruled}
\newfloat{listing}{tb}{lst}{}
\floatname{listing}{Listing}
\usepackage{graphicx}
\usepackage{caption}
\usepackage{subcaption}
\usepackage{amsmath}
\usepackage{amssymb}

\usepackage{booktabs, siunitx} 
\usepackage{makecell}

\title{Zero-Shot Transfer Capabilities of the Sundial Foundation Model for Leaf Area Index Forecasting}
\author{
Peining Zhang\textsuperscript{\rm 1}, 
Hongchen Qin\textsuperscript{\rm 2},
Haochen Zhang\textsuperscript{\rm 1}, 
Ziqi Guo\textsuperscript{\rm 2}, 
Guiling Wang\textsuperscript{\rm 2}, 
Jinbo Bi\textsuperscript{\rm 1}
\thanks{Accepted at the First International Workshop on AI in Agriculture (Agri AI), co-located with AAAI 2026.}
}
\affiliations{
\textsuperscript{\rm 1}School of Computing\\
\textsuperscript{\rm 2}School of Civil and Environmental Engineering\\
University of Connecticut, Storrs, CT 06269, USA

\textrm{peining.zhang, hongchen.qin, haochen.zhang, ziqi.guo, guiling.wang, jinbo.bi@uconn.edu}
}
\begin{document}
    \maketitle

    \begin{abstract}
    This work investigates the zero-shot forecasting capability of time series foundation models for Leaf Area Index (LAI) forecasting in agricultural monitoring. Using the HiQ dataset (U.S., 2000-2022), we systematically compare statistical baselines, a fully supervised LSTM, and the Sundial foundation model under multiple evaluation protocols. We find that Sundial, in the zero-shot setting, can outperform a fully trained LSTM provided that the input context window is sufficiently long—specifically, when covering more than one or two full seasonal cycles. We show that a general-purpose foundation model can surpass specialized supervised models on remote-sensing time series prediction without any task-specific tuning. These results highlight the strong potential of pretrained time series foundation models to serve as effective plug-and-play forecasters in agricultural and environmental applications.
    \end{abstract}

    \begin{links}
    \link{Code}{https://github.com/peiningzhang/sundial-lai}
    \link{Datasets}{https://essd.copernicus.org/articles/16/1601/2024/}
    \end{links}

\section{Introduction}
    Leaf Area Index (LAI) is a critical biophysical variable widely used in agricultural and ecological monitoring, quantifying canopy structure and influencing crop growth. 
    Beyond its central role in agronomic decision-making and yield estimation, accurate LAI forecasting is vital for understanding land-surface feedbacks, as vegetation significantly influences regional climate through biogeophysical processes (e.g., albedo and roughness) and biogeochemical cycles~\cite{alessandri2017multi}. 
    However, practical forecasting is challenged by diverse cropping systems, varying environmental conditions, and limited labeled data. This motivates models that can generalize across locations and years. 
    Furthermore, supervised models trained on heterogeneous datasets tend to learn "global average" patterns and may miss site-specific phenological anomalies without extensive local fine-tuning.

    Recent advances in time series foundation models, such as Sundial~\cite{liu2025sundial}, offer a promising avenue for improving LAI forecasting. Unlike task-specific models, these large, pretrained networks leverage in-context learning, treating historical time series as "prompts" to deduce underlying dynamics. 
    This paradigm has the potential to adapt to new tasks and domains more efficiently than traditional supervised approaches.
    
    An important research question we address is: \textbf{Can a general-purpose, zero-shot time series foundation model match or even surpass a supervised model (e.g., LSTM) specifically trained for LAI prediction?}
    
    This paper makes the following three contributions:
    
\begin{itemize}
    \item We conduct a rigorous zero-shot benchmark of the Sundial foundation model for LAI remote sensing prediction using the comprehensive HiQ dataset.
    \item We demonstrate that, given a sufficiently long historical input context, Sundial (zero-shot) outperforms a fully supervised LSTM. This challenges the conventional assumption that task-specific fine-tuning is necessary for optimal performance and highlights the feasibility of "plug-and-play" forecasting.
    \item We analyze the context window threshold where zero-shot performance overtakes the supervised baseline, and discuss the distinct mechanisms at play: instance-level adaptation in foundation models versus dataset-level generalization in LSTMs.
\end{itemize}
    
    Our findings offer new empirical insights into the potential and limitations of time series foundation models for remote sensing vegetation analysis, and motivate future research into scalable agricultural forecasting.

\section{Related Work}

    \subsection{LAI Time Series Forecasting}
    
    Leaf Area Index (LAI) is a key indicator for monitoring vegetation health, phenology, and long-term ecosystem dynamics.
    Most LAI forecasting approaches treat the signal as a univariate time series characterized by strong seasonality, low-frequency variation, and limited annotated samples~\cite{jiang2010modeling}. Classical methods include harmonic models, autoregressive formulations, and exponential-smoothing variants, which capture seasonal cycles but struggle with nonlinear responses to climate and land-cover variability~\cite{winters1960forecasting}. Recent studies explore data-driven forecasting of vegetation indices, yet robust LAI prediction remains challenging due to heterogeneous temporal patterns and spatial variability across regions~\cite{liu2022estimation}.
    \subsection{Baselines: Statistical and Supervised Models}
    
    Classical LAI forecasting has long relied on statistical time series models such as autoregressive (AR) and exponential-smoothing formulations, which are simple, interpretable, and effective for short-term dynamics. However, these models are limited in their ability to capture nonlinear seasonal patterns and interannual variability. 

    Supervised machine-learning baselines—including random forests~\cite{breiman2001random} and gradient boosting~\cite{friedman2001greedy, ke2017lightgbm}—can incorporate richer auxiliary features when available, but offer only marginal improvements when restricted to univariate LAI histories. Recurrent neural networks, particularly LSTMs, remain strong supervised baselines for vegetation-index forecasting and are widely used in agronomic and ecological applications~\cite{hochreiter1997long, liu2022machine}. Despite their capacity to model nonlinear temporal dependencies, these methods require substantial labeled data and typically exhibit limited gains when training data is scarce or spatially heterogeneous.
    
    Importantly, none of these approaches leverage cross-domain temporal knowledge. As a result, their generalization under long sequences, sparse supervision, or unseen spatial regions remains constrained—highlighting the value of pretrained time series foundation models, which can transfer temporal priors learned from large and diverse corpora.
    
    \subsection{Time Series Foundation Models}
    Large pretrained time series models have recently demonstrated strong zero-shot and few-shot transfer across diverse domains.
    Models such as TimesFM and Sundial learn universal temporal representations from mixed-modality corpora spanning climate, sensor networks, and economic indicators~\citep{das2024decoder, liu2025sundial, ansari2024chronos}.
    These foundation models can leverage broad temporal priors—seasonality, trends, and multi-scale structure—reducing dependence on in-domain training data and enabling robust downstream forecasting.

\section{Data}

    \subsection{HiQ Dataset}
    
    We conduct our experiments using the HiQ dataset~\citep{yan2023hiq}, a large-scale benchmark for Leaf Area Index (LAI) prediction. HiQ reprocesses the satellite MODIS LAI product~\citep{MODIS_MYD15A2H_2015} using a spatiotemporal information compositing algorithm, resulting in improved spatiotemporal consistency and reduced noise. The dataset provides high-quality global LAI time series spanning 2000--2022, with broad geographic coverage across diverse crop types and ecological regions.
    
    In this work, we restrict our analysis to the continental United States (CONUS) subset. Although HiQ offers fully available global coverage, this choice is primarily driven by computational constraints rather than data availability or methodological limitations. We argue that the CONUS domain already encompasses a wide range of vegetation regimes and climatic conditions, making it a representative testbed for evaluating zero-shot LAI forecasting performance.
    
    We use the HiQ product at 5-km spatial resolution, where each pixel contains an LAI time series sampled at an 8-day cadence, resulting in approximately 1,050 time steps per location.

    \subsection{Seasonal snapshots}
\begin{figure}
        \centering
        \includegraphics[width=\linewidth]{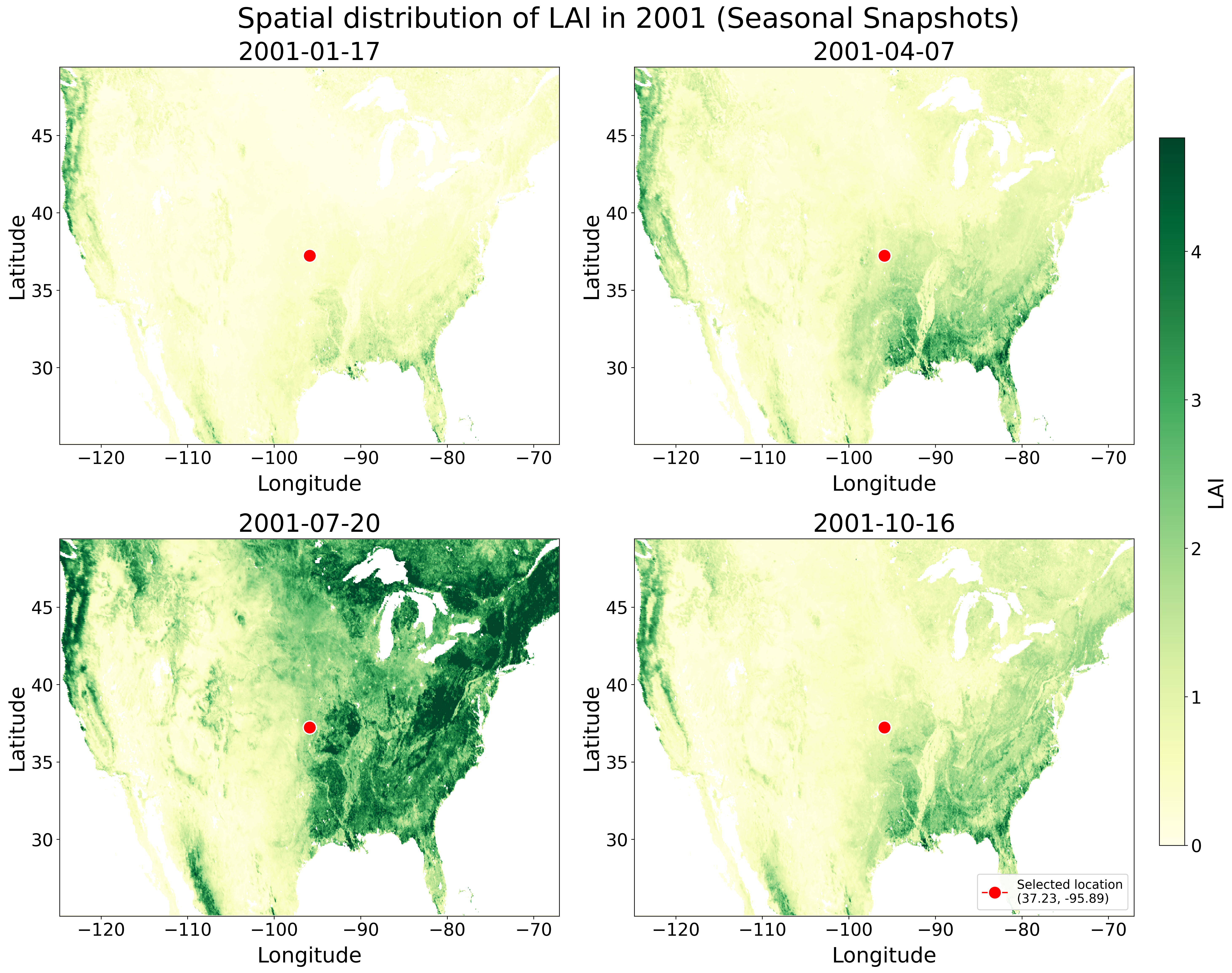}
        \caption{Spatial distribution of LAI across the study domain, shown as four seasonal snapshots from 2001. Higher LAI values (darker green) correspond to denser vegetation.}
        \label{fig:spatial_lai}
    \end{figure}

    The HiQ dataset exhibits substantial spatial and temporal heterogeneity across the continental United States, as illustrated in Figure~\ref{fig:spatial_lai}. The four seasonal snapshots from 2001 reveal pronounced geographic gradients in vegetation greenness, with distinct regional patterns that evolve throughout the year. This spatial variability, combined with the strong seasonal cycles evident in the temporal evolution, presents a challenging forecasting problem that requires models capable of capturing both long-term seasonal patterns and location-specific dynamics. The pronounced seasonal variations highlighted in these snapshots underscore the importance of sufficient context window length—covering multiple full seasonal cycles—for accurate LAI forecasting, which directly relates to our finding that foundation models like Sundial require extended input windows to achieve optimal zero-shot performance.
    
    \subsection{Temporal Dynamics}
\begin{figure}
    \centering
    \includegraphics[width=1\linewidth]{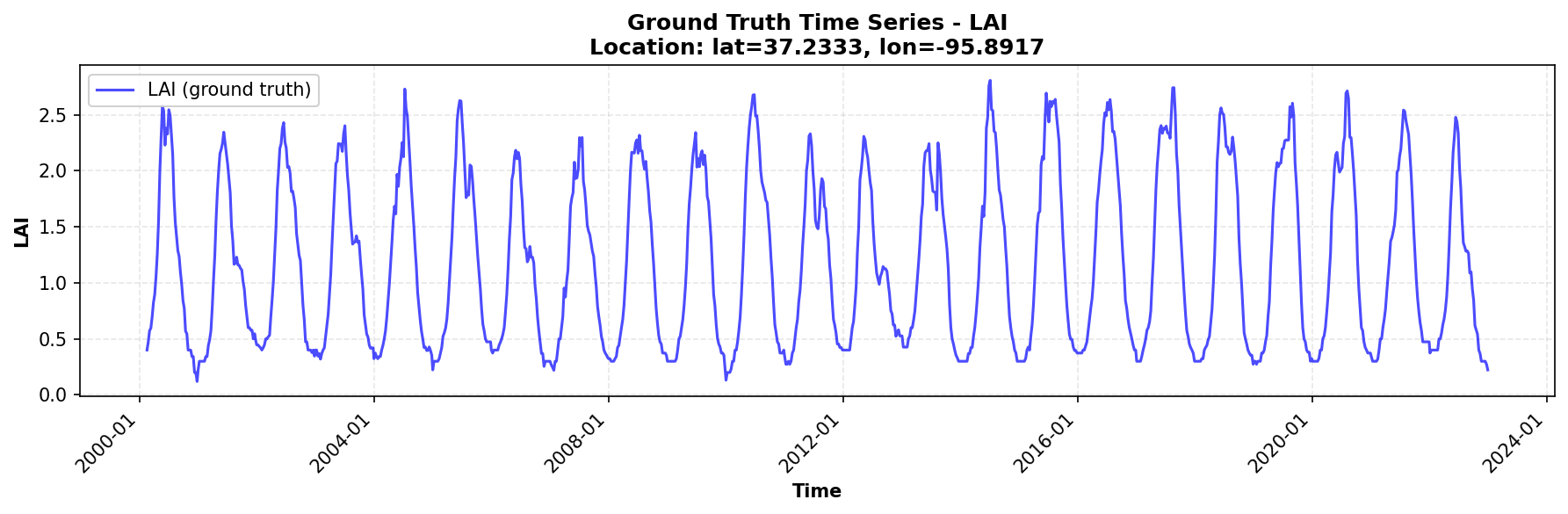}
    \caption{Temporal LAI evolution at the representative center pixel (red dot in Fig.~\ref{fig:spatial_lai}). The series shows clear seasonal cycles and interannual variability.}
    \label{fig:ground_truth}
    \end{figure}

    To better understand the temporal behavior of the vegetation data, we examine the LAI time series of a representative pixel, referred to as the "center location." This specific location is geographically highlighted by a red dot in Figure \ref{fig:spatial_lai}. Figure \ref{fig:ground_truth} presents the complete temporal profile for this location, where the horizontal axis represents the date and the vertical axis indicates the LAI value. The series exhibits pronounced seasonal periodicity, reflecting the natural phenological cycle of vegetation growth and senescence typical for this region.
    
    \subsection{Preprocessing and Quality Control}
     To ensure consistency, all LAI time series undergo per-pixel normalization. For supervised models (e.g., LSTM), we normalize each pixel's time series using its mean and standard deviation computed from valid (non-missing) observations in the training period. Missing or masked values, common in satellite products due to clouds or sensor issues, are handled in two stages. For most models, missing values are either explicitly masked out during loss computation, or---where required---are imputed with linear interpolation over time. Cells with a very high fraction of missing data (i.e. $>0.1$) are excluded from the training and evaluation splits.

\section{Methods}

    \subsection{Baseline Models}

    We consider a suite of established baselines for LAI forecasting:

    \begin{itemize}
        \item \textbf{Last Value}: Predicts the most recent observed LAI from the context window as the forecast for all future steps.
        \item \textbf{Mean(k)}: Uses the mean of the last $k$ observed LAI values as the prediction for all horizons.
        \item \textbf{Trend}: Extrapolates a fitted linear trend from the previous $k$ time points to predict future LAI values~\cite{jensen2008discrete}.
        \item \textbf{ARIMA}: For each test window, we fit an ARIMA$(2,1,0)$ model~\cite{box1970distribution, seabold2010statsmodels} to the observed LAI values to forecast the next value. ARIMA is chosen as a simple yet reasonable baseline for univariate time series; model fitting uses state-space estimation.
        \item \textbf{LSTM}: A supervised Long Short-Term Memory model~\citep{hochreiter1997long} trained on the entire training set to capture nonlinear dependencies in LAI time series. Our LSTM architecture consists of two stacked LSTM layers with 64 hidden units each, followed by a fully connected output layer. The model is trained using the Adam~\cite{kingma2014adam} optimizer with a learning rate of 0.001, batch size of 256, and early stopping based on validation loss.
    \end{itemize}
    \begin{figure}[t]
        \centering
        \includegraphics[width=\linewidth]{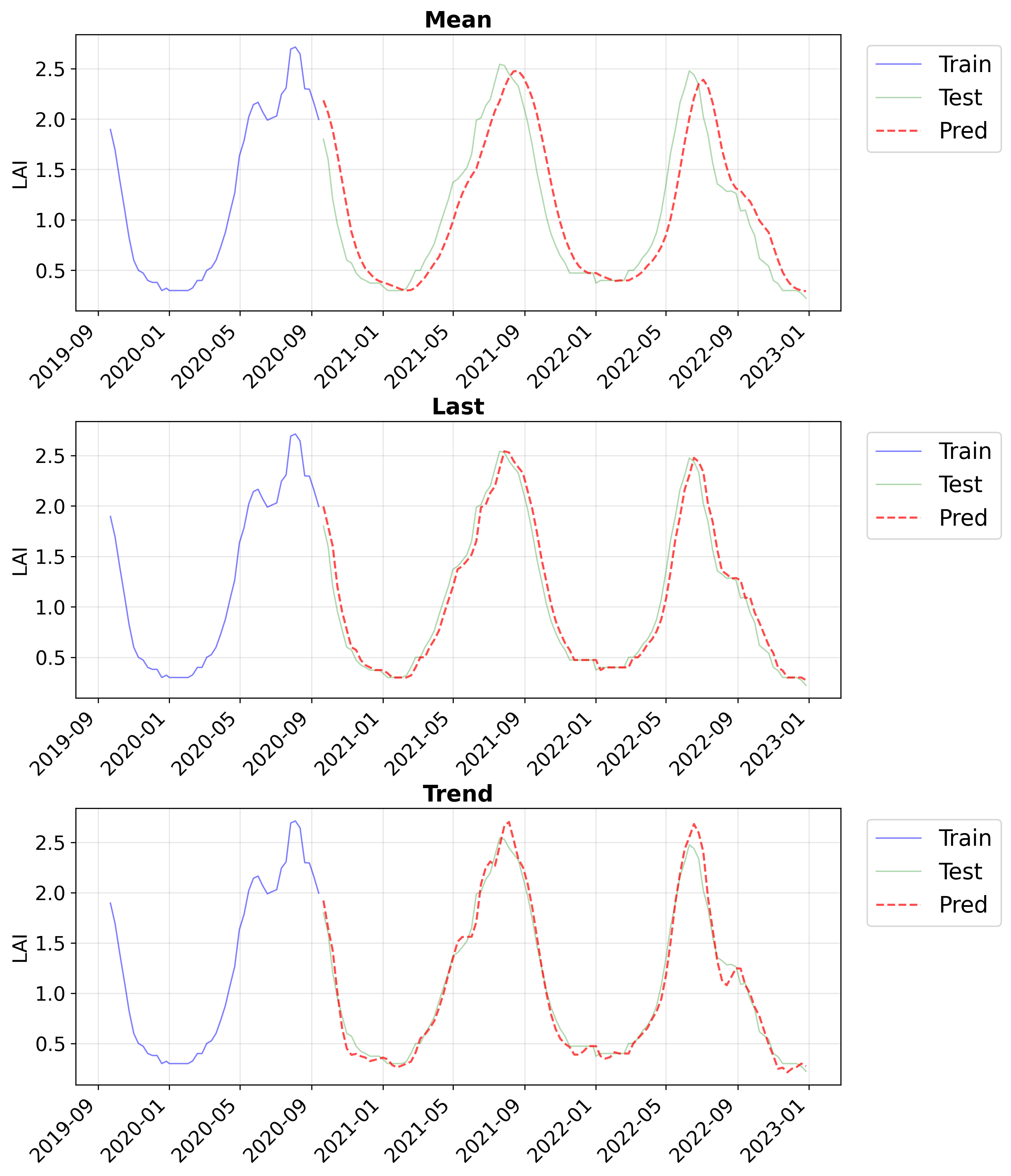}
        \caption{
            \textbf{Performance of simple statistical baselines on the final 150 time steps of a representative LAI pixel.}
            The first 46 time steps show ground-truth LAI, followed by 104 forecasted steps.
            The \emph{Mean} baseline underestimates seasonal amplitude, the \emph{Last} baseline lags the seasonal cycle, and the \emph{Trend} baseline tracks short-term tendencies but fails at seasonal turning points.
        }
        \label{fig:baseline_predictions}
    \end{figure}

    \subsection{Forecasting Protocol}
    We adopt a unified fixed-window forecasting protocol across all models. For each pixel, we generate samples using a sliding window with input length $T_\text{in}$ and forecast horizon $H$.     
    To evaluate scalability, we test multiple window sizes $T_\text{in} \in \{32, 64, 128, 256, 512, 1024\}$ and forecast horizons $H \in \{1, 4, 8, 12\}$ (corresponding to 8, 32, 64, and 96 days ahead, respectively, given the 8-day temporal resolution). 
    Evaluation follows a strict forward-in-time split: models are trained (or conditioned) on historical segments and tested on the most recent unseen data. 
    For the supervised baselines (e.g., LSTM), parameters are optimized by minimizing the Mean Squared Error (MSE) between predictions and ground truth. In contrast, the foundation model operates in strict inference-only mode without task-specific loss optimization.

    \subsection{Zero-Shot Inference with Sundial}
    We formulate the LAI forecasting task as a conditional generation problem. 
    Instead of fine-tuning model weights, we leverage the pre-trained capabilities of Sundial directly~\cite{liu2025sundial}. 
    The historical input window $X \in \mathbb{R}^{T_\text{in}}$ serves as a continuous \textit{context prompt}, analogous to a text prompt in Large Language Models (LLMs). 
    The model processes this prompt to deduce the underlying phenological dynamics (e.g., seasonality and trend) and generates the future sequence. 
    This approach relies entirely on the model's in-context learning ability acquired during pre-training, enabling a strictly zero-shot evaluation. We use the publicly available Sundial model checkpoint without any modifications or fine-tuning. For inference, we generate forecasts autoregressively for multi-step horizons.


\section{Experiments}
    \subsection{Evaluation Metrics}
    We evaluate model performance using standard error- and scale-aware regression metrics:
    
    \begin{itemize}
        \item \textbf{MAE}: Mean Absolute Error
        \item \textbf{RMSE}: Root Mean Square Error
        \item \textbf{R\textsuperscript{2}}: Coefficient of Determination
        \item \textbf{MAPE}: Mean Absolute Percentage Error
        \item \textbf{CVRMSE}: Coefficient of Variation of RMSE
    \end{itemize}
    
    Metrics are aggregated over 100 randomly sampled locations to ensure computational efficiency. With approximately 100 rolling test windows per pixel, each reported metric represents an average across $\sim 10{,}000 \times H$ prediction-ground truth pairs, where $H$ is the forecast horizon.
    
\begin{figure}[!t]
        \centering
        \includegraphics[width=\linewidth]{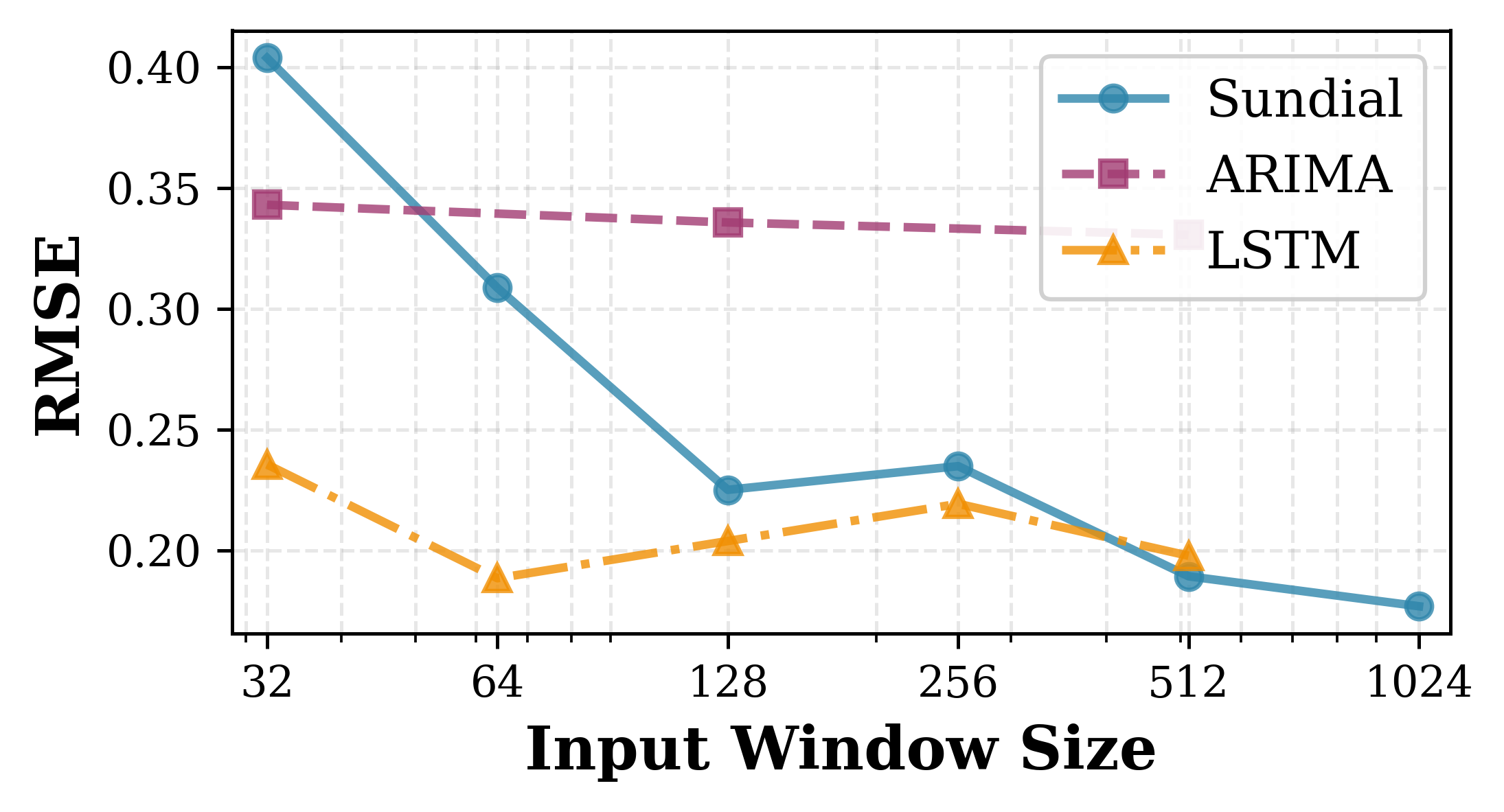}
        \caption{\textbf{Effect of input window size ($T_{\text{in}}$) on forecasting accuracy at horizon $H=4$.}
            RMSE comparison across different window sizes. Larger input windows consistently improve Sundial performance,
            while ARIMA and LSTM saturate at moderate window lengths. Metrics are averaged per pixel.}
        \label{fig:window-size-effect}
    \end{figure}
    \begin{figure}
        \centering
        \includegraphics[width=1\linewidth]{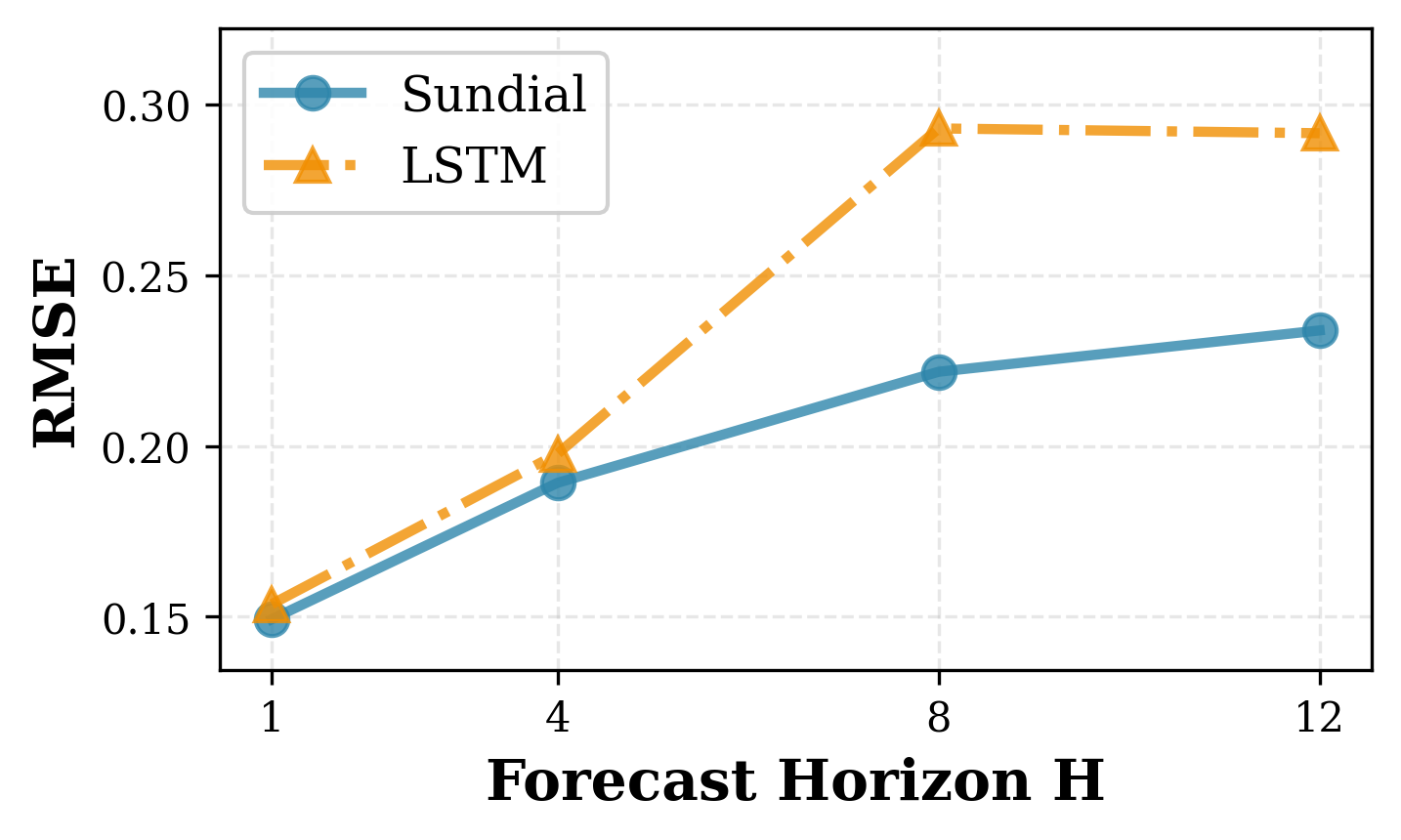}
        \caption{\textbf{Effect of forecasting horizon ($H$) on forecasting accuracy.} Multi-horizon forecasting comparison between Sundial and the supervised LSTM at a fixed context window ($T_\text{in}$ = 512). The x-axis denotes the forecasting horizon H $\in$ \{1, 4, 8, 12\}, and the y-axis shows the corresponding average RMSE. Although both models degrade as $H$ increases, the zero-shot Sundial model exhibits a substantially slower performance deterioration, outperforming the supervised LSTM consistently for all $H \geq 4$. This demonstrates that Sundial retains long-range temporal structure more effectively, making it particularly suitable for extended-horizon agricultural forecasting.}
        \label{fig:forecast-length-effect}
    \end{figure}
    
    \subsection{Rule-based Baselines Results}
    
    To contextualize the difficulty of LAI forecasting, we first evaluate three simple rule-based baselines: Mean, Last Value, and Trend.
    As shown in Fig.~\ref{fig:baseline_predictions}, these heuristics struggle to follow the seasonal dynamics of LAI, with Mean lagging transitions, Last Value producing phase-shifted predictions, and Trend failing near turning points.
    These behaviors illustrate the limitations of simple historical extrapolation and motivate the need for more expressive forecasting models.

    \begin{table*}[!h]
    \centering
    \small
    \begin{tabular}{
    l
    c
    S[table-format=1.3]
    S[table-format=1.3]
    S[table-format=2.2]
    S[table-format=2.2]
    S[table-format=2.2]
    }
    \toprule
    \textbf{Model} & \textbf{Window} & {\textbf{MAE} ($\downarrow$)} & {\textbf{RMSE} ($\downarrow$)} & {\textbf{$R^2$} (\%) ($\uparrow$)} & {\textbf{MAPE} (\%) ($\downarrow$)} & {\textbf{CVRMSE} (\%) ($\downarrow$)} \\
    \midrule
    \multicolumn{7}{l}{\textbf{Statistical Baselines}} \\
    Mean   & 4  & 0.274 & 0.375 & 72.6 & 36.37 & 32.90 \\
    Last   & 4  & 0.144 & 0.210 & 87.6 & 17.88 & 19.30 \\
    Trend  & 4  & 0.143 & 0.210 & 85.7 & 20.84 & 19.36 \\
    \midrule
    
    \multicolumn{7}{l}{\textbf{ARIMA}} \\
    ARIMA(2,1,0) & 32     & 0.138 & 0.204 & 87.9 & 17.46 & 18.80 \\
    ARIMA(2,1,0) & 128    & 0.134 & 0.197 & 88.6 & 17.31 & 18.18 \\
    ARIMA(2,1,0) & 512    & 0.134 & 0.196 & 88.7 & 17.15 & 18.08 \\
    \midrule
    
    \multicolumn{7}{l}{\textbf{LSTM (Supervised Baseline)}} \\
    LSTM & 32   & 0.107 & 0.155 & 96.82 & 20.20 & 17.73 \\
    LSTM & 64   & 0.097 & 0.140 & \textbf{97.34} & \textbf{18.83} & 15.78 \\
    LSTM & 128  & 0.097 & 0.141 & 97.33 & 19.53 & 15.97 \\
    LSTM & 256  & 0.105 & 0.148 & 97.23 & 22.86 & 18.56 \\
    LSTM & 512  & 0.099 & 0.143 & 97.26 & 20.44 & 16.45 \\
    \midrule
    
    \multicolumn{7}{l}{\textbf{Sundial (Plug-and-Play)}} \\
    Sundial & 32    & 0.154 & 0.209 & 94.12 & 33.66 & 22.70 \\
    Sundial & 64    & 0.132 & 0.185 & 95.18 & 26.32 & 19.97 \\
    Sundial & 128   & 0.107 & 0.158 & 96.64 & 20.76 & 17.49 \\
    Sundial & 256   & 0.112 & 0.162 & 96.41 & 21.80 & 17.71 \\
    Sundial & 512   & 0.099 & 0.145 & 97.15 & 20.29 & 15.94 \\
    Sundial & 1024  & \textbf{0.096} & \textbf{0.140} & 97.32 & 19.15 & \textbf{15.44} \\
    \bottomrule
    \end{tabular}
    \caption{
    \textbf{Performance benchmark for one-step-ahead LAI forecasting ($H=1$).}
    All entries reflect the updated LSTM and Sundial results across input window sizes ($T_{\text{in}}$). Best values within each model category are \textbf{bolded}.}
    \label{tab:main-results}
    \end{table*}

\begin{table}[ht]
\centering
\small
\begin{tabular}{lccccc}
\toprule
\textbf{\makecell{Missing \\ ratio}} & \multicolumn{5}{c}{\textbf{RMSE increase over baseline (\%)}} \\
\cmidrule(lr){2-6}
 & \textbf{1024} & \textbf{512} & \textbf{256} & \textbf{128} & \textbf{64} \\
\midrule
5\%  & 8.2\%  & 5.1\%  & 1.5\%  & 7.0\%  & 0.8\%  \\
10\% & 10.2\% & 10.9\% & 2.1\%  & 8.9\%  & 4.2\%  \\
20\% & 25.6\% & 21.5\% & 12.8\% & 22.1\% & 7.5\%  \\
40\% & 57.1\% & 53.7\% & 42.3\% & 54.9\% & 36.7\% \\

\bottomrule
\end{tabular}
\caption{\textbf{Sundial robustness under random missing observations at forecast horizon $H=1$.} Entries are relative RMSE increases (\%) compared to the no-missing-data baseline.}
\label{tab:sundial-missing-robustness}
\end{table}

    
\begin{table}[h!]
\centering
\small
\begin{tabular}{lcccccc}
\toprule
\textbf{Model} & \textbf{H\textbackslash$T_{\text{in}}$} &
\textbf{1024} & \textbf{512} & \textbf{256} &
\textbf{128} & \textbf{64}  \\
\midrule
\multicolumn{7}{l}{\textbf{LSTM (Supervised Baseline) — RMSE ($\downarrow$)}} \\
 & 1  & --     & 0.143& 0.157 & 0.150 & 0.149  \\
 & 4  & --     & 0.198 & 0.219 & 0.204 & 0.188  \\
 & 8  & --     & 0.251 & 0.293 & 0.238 & 0.247  \\
 & 12 & --     & 0.269 & 0.292 & 0.265 & 0.261 \\
\midrule
\multicolumn{7}{l}{\textbf{Sundial (Zero-Shot) — RMSE ($\downarrow$)}} \\
 & 1  & 0.140& 0.145& 0.162& 0.158& 0.185\\
 & 4  & 0.177 & 0.189 & 0.235 & 0.225 & 0.309 \\
 & 8  & 0.202 & 0.222 & 0.290 & 0.283 & 0.433  \\
 & 12 & 0.210 & 0.234 & 0.316 & 0.303 & 0.532  \\
\midrule
\multicolumn{7}{l}{\textbf{LSTM - Sundial ($\Delta$ RMSE)}} \\
 & 1  & --    & -0.002 & -0.019 & -0.022 & -0.052  \\
 & 4  & --    &  0.009 & -0.016 & -0.021 & -0.120  \\
 & 8  & --    &  0.030 &  0.003 & -0.045 & -0.186  \\
 & 12 & --    &  0.035 & -0.025 & -0.038 & -0.271  \\
\bottomrule
\end{tabular}
    \caption{
    \textbf{Unified comparison of multi-horizon ($H$) and multi-window ($T_{\text{in}}$) forecasting performance.}
    Cells contain RMSE (lower is better).
    Sundial exhibits monotonic improvement with longer context windows and consistently outperforms LSTM (Supervised Baseline) for $T_{\text{in}} \ge 512$, especially at longer horizons ($H \ge 4$). In the third section, positive values indicate Sundial achieves lower RMSE (better performance).
    }
\label{tab:unified-matrix}
\end{table}

    \subsection{Model Results}    
    \subsubsection{One-Step-Ahead Forecasting Performance}
    Table~\ref{tab:main-results} reports the one-step-ahead forecasting results ($H=1$).  
    Classical statistical baselines and ARIMA establish a lower-bound benchmark, but their errors remain noticeably higher than those of learning-based methods.
    
    Among supervised models, the LSTM achieves its best performance at relatively short windows, with the lowest RMSE (0.140) and highest $R^2$ (97.34\%) obtained at $T_{\text{in}} = 64$. Performance degrades when the window becomes either too small or too large, reflecting the model’s limited ability to exploit extended temporal context.
    
    In contrast, as illustrated in Figure~\ref{fig:window-size-effect}, \textbf{Sundial} benefits steadily from longer input windows. Its accuracy improves monotonically with window size and reaches its best performance at $T_{\text{in}} = 1024$, achieving the lowest MAE (0.096). Despite requiring no task-specific training, Sundial matches or surpasses the supervised baseline when provided with sufficient historical context, highlighting its strong plug-and-play capability for pixel-level LAI forecasting.

    \subsubsection{Robustness to Missing Data}
    To assess the robustness of \textbf{Sundial} to missing observations, we randomly mask a fraction of values in the input context at rates of 5\%, 10\%, and 20\% without interpolation. 
    Notably, Sundial is equipped with a native pipeline for handling NaN inputs.
    We report performance at forecast horizon $H=1$ as the relative RMSE increase (\%) compared to the no-missing-data baseline (Table~\ref{tab:sundial-missing-robustness}).

    \subsubsection{Multi-step Forecasting Across Longer Horizons}
    We evaluate multi-horizon forecasting for $H\in\{1,4,8,12\}$ at a representative context $T_{\text{in}}=512$ (see Figure~\ref{fig:forecast-length-effect}). Both models' errors increase with $H$, but Sundial's degradation is substantially slower: for $H\ge4$ Sundial consistently outperforms the supervised LSTM despite remaining zero-shot. Although the per-pixel RMSE varies significantly across the dataset (high standard deviation due to spatial heterogeneity), a paired t-test confirms that Sundial's performance improvement over LSTM at this horizon ($H=4$) is highly statistically significant ($p < 0.001$, $N \approx 10,000$). This pattern reinforces that, given sufficient context, the foundation model better preserves pixel-specific temporal structure for extended horizons.

     \subsubsection{Scalability in Multi-Step Forecasting}
    The unified results in Table~\ref{tab:unified-matrix} reveal a clear divergence in how the two models respond to longer input windows and forecasting horizons. The supervised LSTM reaches its best accuracy at short windows (e.g., $T_{\text{in}}=64$) but consistently degrades as the window grows, reflecting its limited ability to retain long-range temporal dependencies.
    
    In contrast, Sundial benefits steadily from additional historical context. Its RMSE improves monotonically with increasing window size and eventually surpasses the LSTM baseline once $T_{\text{in}} \ge 512$, a trend that becomes more pronounced at longer horizons ($H \ge 4$). This behavior highlights Sundial’s capacity to leverage extended seasonal information and maintain stability under multi-step forecasting, a key requirement for agricultural monitoring.

\section*{Discussion}

    Our experiments reveal a compelling crossover: while the supervised LSTM dominates in short-context regimes, the zero-shot Sundial foundation model significantly outperforms it given sufficient history. This finding highlights the trade-off between two distinct learning paradigms.
    
    \paragraph{Global Priors vs. Local Adaptation.} 
    The performance disparity observed across window sizes stems from how each model utilizes information. The supervised LSTM relies on \textbf{dataset-level generalization}, excelling at $T_\text{in} \le 64$ by leveraging learned "global average" priors to extrapolate incomplete seasonal cycles. In contrast, Sundial relies on \textbf{instance-level in-context learning}. While it struggles with short prompts, in long-context regimes ($T_\text{in} \ge 512$), the extended window enables it to capture \textbf{unique pixel-specific signatures}—such as specific drought responses or phenological shifts—that the generalized LSTM tends to smooth over. Effectively, Sundial builds a temporary, local model for each pixel in real-time, surpassing the static global template of the LSTM.
    
    \paragraph{Towards "Plug-and-Play" Remote Sensing.} 
    These results suggest a potential paradigm shift. We demonstrate that a pre-trained foundation model can be deployed "off-the-shelf" to achieve state-of-the-art forecasting accuracy. This \textbf{"plug-and-play"} capability offers a scalable alternative to the resource-intensive "collect, annotate, and train" workflow, particularly valuable for agricultural monitoring in regions where training data is scarce or rapid deployment is required.
    
    \paragraph{Limitations and Future Directions.} 
    While promising, several limitations warrant discussion. First, the requirement for long context windows ($T_\text{in} \ge 512$, approx. 11 years) may limit applicability in regions with sparse historical records. Second, our evaluation focuses on univariate forecasting; incorporating multi-modal inputs (e.g., climate variables, soil properties) could further enhance accuracy. Third, the inference computational cost of foundation models remains higher than lightweight baselines. Future work will focus on exploring few-shot fine-tuning to reduce context requirements and expanding geographic validation to assess global transferability.

\section*{Conclusion and Future Work}
    In this work, we conducted a systematic evaluation of the zero-shot capabilities of the \textbf{Sundial} time series foundation model for remote-sensing LAI forecasting. We found that, when provided with a sufficiently long historical context window, Sundial's zero-shot performance \textbf{can surpass} that of a supervised \textbf{LSTM} model specifically trained for the task, challenging the conventional view that task-specific training guarantees superior performance. Our findings suggest that these general-purpose models already encode powerful temporal patterns and vegetation phenology, enabling them to serve as highly efficient and powerful \textbf{plug-and-play} forecasting tools without the need for any task-specific training.

    Looking ahead, we plan several extensions to further enhance model applicability and robustness:
    \begin{itemize}
        \item Incorporating additional climatological drivers (e.g., temperature, precipitation) as input features to improve forecasts under varying environmental conditions;
        \item Exploring multi-variate modeling to enable the joint modeling of multiple vegetation indices and ecosystem variables for more comprehensive ecological status prediction; and
        \item Conducting broader spatial transfer experiments at larger geographic scales to rigorously evaluate model generalization, thereby supporting global vegetation monitoring.
    \end{itemize}

\section*{Acknowledgments}
    This work was partially supported by the U.S. National Science Foundation (NSF) under Grant AGS-2426655. We thank the anonymous reviewers for their constructive comments and suggestions. Any opinions, findings, and conclusions or recommendations expressed in this material are those of the authors and do not necessarily reflect the views of the NSF.

    \bibliography{aaai2026}
\end{document}